# Exploring Human's Gender Perception and Bias toward Non-Humanoid Robots

Mahya Ramezani and Jose Luis Sanchez-Lopez

**Abstract**

In this study, we investigate the human perception of gender and bias toward non-humanoid robots. As robots increasingly integrate into various sectors beyond industry, it is essential to understand how humans engage with non-humanoid robotic forms. This research focuses on the role of anthropomorphic cues, including gender signals, in influencing human-robot interaction and user acceptance of non-humanoid robots. Through three surveys, we analyze how design elements—such as physical appearance, voice modulation, and behavioral attributes—affect gender perception and task suitability. Our findings demonstrate that even non-humanoid robots like Spot, Mini-Cheetah, and drones are subject to gender attribution based on anthropomorphic features, affecting their perceived roles and operational trustworthiness. The results underscore the importance of balancing design elements to optimize both functional efficiency and user relatability, particularly in critical contexts.

## I. Introduction

In this fast-developing technological world of today, robots are gradually becoming part of several facets of human society. They cannot be only restricted to the industries but now can be found extensively in health care services, search and rescue operations, environmental monitoring, and even social companionship[1]. Robots such as drones and exploration robots have demonstrated superior performance in a variety of tasks, often surpassing what humanoid robots can achieve [2]. Drones are used in aerial mapping and surveillance[3, 4]; quadrupedal robots can traverse rough terrains, thus being invaluable in search and rescue operations and the agricultural industry[5]. All these point to the need to understand Human-Robot Interaction (HRI) and how humans perceive and interact with non-humanoid robots[6].

Recent research has illuminated the innate human tendency to anthropomorphize, a phenomenon wherein human-like qualities are ascribed to inanimate objects and non-human entities, influencing our interactions with them [7]. Features of robots in terms of HRI with physical and behavioral attributes serve an invaluable function in creating human engagement [8]. Some selected features of anthropomorphism and some aspects of social behavior should also be embedded into robot design to enable their acceptance. Designing and endowing robots according to human expectation can drastically enhance the acceptance and performance of robots [9, 10].

However, achieving the right balance in robot design is crucial. Excessive anthropomorphism in robots increases the likelihood of invoking the uncanny valley [11] effect, which research has shown to induce discomfort and unease among users[12]. Conversely, robots that lack sufficient humanoid features may struggle to engage users effectively. Additionally, non-humanoid robots offer unique advantages in design and functionality that cannot be achieved with humanoid robots based on their applications. This challenge is particularly relevant for non-humanoid robots, where explicit human-like features are often absent. Based on our research Incorporating anthropomorphic elements, such as gender cues, into non-humanoid robots can bridge this gap and have the potential to enhance relatability and facilitate greater acceptance. Understanding human perception of non-humanoid robots and determining how humans identify non-humanoid robots is crucial for achieving this goal. By investigating how people perceive non-humanoid robots and how gender cues and other anthropomorphic elements can be integrated into non-humanoid robot design, this research aims to enhance HRI and develop robots that are more effective, trustworthy, and relatable.

### 1) *Anthropomorphism in Robots*

Anthropomorphism is the attribution of human characteristics, emotions, and intentions to other entities that are not human. It creates attraction and acceptance among users; therefore, it is one of the most important elements that influences HRI. The level of anthropomorphism in robot design determines how humans would act in response to the robots.

For instance, the service robots in the hospitality industry are increasingly being designed to take on human-like features to make customer experience better, reduce labor cost, and build touchless systems. Humanized service robots, like Sophia [13], a social humanoid robot is designed to learn and adapt to human behavior and work with humans, making her a popular figure for demonstrations and interactions in various industries. Sophia's advanced AI and ability to express a range of emotions make her an engaging tool for human-robot interaction studies. In healthcare, humanoid robots like Moxi [14] assist with logistics and routine tasks in hospitals. Moxi is designed to interact with staff and patients, to provide a friendly and effective service to help reduce the workload of healthcare professionals. In the retail sector, robots like SoftBank's NAO [15] are used to greet customers, provide information, and even conduct transactions. NAO's humanoid design and ability to interact using natural language processing make it an asset for enhancing customer experience and operational efficiency. These features and characteristics of these humanoid robots, to be more effective, are endowed with more anthropomorphism, which by default enhances their operational effectiveness and user engagement, ergo user satisfaction. This is why anthropomorphism has always been a crucial factor in robot designing.

For non-humanoid robots, which often lack the obvious human form, the introduction of anthropomorphic cues such as voice modulation or simplified facial features can lead to better acceptance and more intuitive interaction for users[16, 17]. For example, the effectiveness of non-humanoid robots in service roles was enhanced by tailoring their design to match the tasks they perform, ensuring that their appearance supports their functional roles without leading to unrealistic expectations from users[18].

### 2) *Gender Schemas and Stereotypes*

Gender schemas[19] are cognitive structures that organize knowledge about gender and guide responses to gendered information; they are deeply integrated into social cognition and affect our perceptions and interactions with others. These schemas manifest in gender stereotypes, which shape our expectations of certain social groups and roles, such as the perception of service roles as 'female-oriented' tasks[20]. Such stereotypes are not only prevalent in societal roles but also deeply embedded in the design of both humanoid and non-humanoid robots.

In the robot design, humanoid robots often feature gendered characteristics such as body shape, facial features, and voice modulation that align with traditional male or female attributes, enhancing relatability and societal acceptance[21]. The adjectives considered for gendering in research often include 'assertive', 'aggressive' for masculine traits[22], and 'nurturing', 'submissive' for feminine traits, influencing the design choices like body color, hair length, and decorative elements which shape gender perceptions[23, 24]. Non-humanoid robots, while lacking human form, still convey gender through names or voices that match roles traditionally associated with specific genders [17], like robots in domestic settings with female names or voices, reinforcing stereotypes of women in caregiving roles[25]. This bias is exemplified in the design of virtual assistants like Amazon's Alexa and Google's Assistant, which are endowed with female identities to evoke a sense of care and approachability[26].

Research shows that these design choices are not superficial; they influence user interactions. Studies by Eyssel and Hegel demonstrate that people attribute gender to robots based on appearance and voice pitch, impacting perceptions and interactions[27]. Robots with feminine features or nurturing roles are often perceived as female, while those with masculine features or authority roles are seen as male. This gendering affects user trust, acceptance, and satisfaction, manifesting in preferences for female robots in caregiving roles and male robots in security roles[25]. Further studies underscore the implications of these perceptions.

Research indicates that users often find robots of the opposite gender more credible and engaging, influencing their emotional responses and behaviors[28]. This effect highlights the impact of gender attribution on robot design and use, potentially leading to biases and affecting human relationships by altering social interactions and empathy levels. For example, Mutlu et al. found that robot appearance and behavior influence acceptance in workplace settings[29], while studies by Kim et al. and Li et al.[30, 31] assess user perceptions of robot utility and safety, noting that friendly and inviting designs enhance robot approachability.

Understanding these dynamics is crucial for designing robots that effectively engage users and meet their expectations, as explained by Bem's gender schema theory[32], which discusses how cognitive structures about gender shape behaviors and perceptions in technological interactions.

### *3) Gaps in Our Knowledge About Gender Attribution to non-humanoid robots*

Although anthropomorphic and gendered aspects of humanoid robots have been the subject of numerous studies, non-humanoid robot research is still in its infancy. Non-humanoid robots present unique design challenges and opportunities that warrant distinct consideration. Unlike their humanoid counterparts, non-humanoid robots such as drones or quadruped robots often operate without clear anthropomorphic cues, which can alter user perceptions and interactions[33]. This difference underscores the importance of our research into how anthropomorphic elements can be applied to non-humanoid robots to enhance their functionality and societal acceptance. This gap draws attention to an important field of research: how does the design of non-humanoid robots affect user interactions when anthropomorphism is incorporated? Does it use the same gender preconceptions as humanoid robots when designing non-humanoid robots in HRI contexts? Knowing this could help determine whether or not gendering non-humanoid robots improves their usability or acceptance by users in a variety of roles.

Furthermore, while surface features and tasks have received the majority of research attention when it comes to humanoid robots, roles and behaviors can also indicate gender, particularly when gender schemas are not incorporated into the design. Gender roles are often gender-specific in many cultures[34].

Additionally, regarding humanoid robots, most studies have focused on surface features and tasks, but gender may also be conveyed through roles and behaviors, especially in the absence of gender schemas in design. For instance, in the Middle East, men are usually connected with sports and socializing, while women are often associated with childcare and housekeeping[35]. It is important to know if non-humanoid robots can be gendered and enculturated like humans.

Finally, it is important to explore whether adding humanized features can improve the relationship between non-humanoid robots and humans. There is limited research on gender differences and individual differences in gendering robots, particularly non-humanoid ones. Furthermore, understanding how these gendered behavioral actions and design features impact efficiency, trust, and performance in real-world missions is essential.

Based on this, our research aims to investigate human perception of the gender of non-humanoid robots using both design and behavioral analysis. We seek to understand how non-humanoid robots are perceived, determining whether surface features alone are significant or if behavioral roles also play a crucial part. We define gender as a spectrum of masculinity and femininity rather than as assigned sex, viewing it as a continuum from 1 to 5.

Given this, we pose the following research questions:

- What are public perceptions of non-humanoid gender based on assigned gendered adjectives and traditional roles?
- How do design elements related to gender cues in non-humanoid robots influence perceptions of anthropomorphism and interaction dynamics in HRI?
- Can genderization in non-humanoid robots enhance effectiveness, trust, and relatability in interactions?
- What are public perceptions and preferences regarding roles and characteristics in non-humanoid robots?

In this research, we selected the Spot and, Mini-Cheetah and Mavic-2 pro, because of their unique designs, which lack explicit human-like features, providing an ideal basis for examining how humans might attribute gender to robots not through humanoid form but via movement, shape, or presumed roles. Given their typical absence of anthropomorphic characteristics in these robots and their general perception as utilitarian devices, they provide a contrast. These choices help us understand gender perceptions in technology for various social uses, guiding the design and training of non-humanoid robots to improve their social acceptability.

To address these research questions, we conducted three surveys. The first survey detect human perceptions of non-humanoid robots. This included techniques such as implicit association tests and analyzing user reactions to various robot actions without explicitly mentioning gender. The aim was to gather unbiased data on how people perceive gender in robots based on their shape and behavioral and roles.

The second survey aimed to investigate the effects of humanizing non-humanoid robots by adding anthropomorphic and gender-stereotyped attributes, based on insights from the literature and the first survey. Design elements such as aspect ratio, head shape, and curves were modified, and facial features added to enhance anthropomorphism. Gender-specific cues, such as voice modulation, were also incorporated. The goal was to assess how different levels of humanization, particularly gender attributes, influence perceptions and behaviors in HRI, with a focus on potential differences in male and female responses.

The third survey specifically focused on the impact of gender cues in drone design on robot acceptance and role assignment in real world mission scenario for its importance to required human robot interaction for mission success[36]. The third study investigated the effects of anthropomorphic design elements, such as gender-specific vocalizations, gender stereotype in movementespecially about the speed, and gendered physical appearances, on human perceptions of safety, comfort, and interaction politeness. The objective was to understand how these design attributes impact the operational effectiveness of drones in SAR tasks and their acceptance by humans. We should note that, in all surveys, participants were assured of the confidentiality and anonymity of their responses. They were also informed that their participation was entirely voluntary and that they could withdraw from the study at any time without any consequences.

## II. Study 1: Examination of Gender Attribution in Non-Humanoid Robots

The primary aim of this study is to examine how individuals perceive non-humanoid robots, focusing specifically on gender attributes. We seek to determine whether people attribute human-like qualities to non-humanoid robots and the extent to which they assign gender-specific attributes. Additionally, we aim to explore the relationship between gender, assigned attributes, and efficiency in various roles for non-humanoid robots in this study. We are also interested in understanding the connection between the robots' behavioral and physical attributes and the roles assigned to them by people, as well as how these factors influence perceptions of the non-humanoid robots' gender. This investigation will inform not only design considerations but also planning and training strategies, ensuring robots can interact with people with behaviors appropriate for their designated tasks.

### *A. Pilot study*

Before the main survey, a pilot study was conducted at the University of Luxembourg in person to explore the feasibility of directly gauging human perceptions of non-humanoid robot gender and to refine the hypotheses, stimuli, and parameters for the main study. The study involved 30 participants, aged between 25 to 55 years, with a gender distribution of 53% male and 47% female. Participants rated their perceived gender of each robot on a scale from 1 (feminine) to 5 (masculine), with 3 indicating a neutral perception.

The Spot robot was perceived as more masculine, with an average score of 3.72 SD=0.85, whereas the Mini-Cheetah robot was seen as more feminine, with an average score of 3.08 SD=1.10 and drone with an average score 3.2 SD=1.05 These results underscore the human tendency to assign gender to robots, even non-humanoid ones. Additionally, participants' ratings of adjectives related to male or female associations showed low standard deviations, indicating a consensus that aligns with societal norms and gender stereotypes.

The study also revealed some participants' reluctance to directly answer questions about robot gender, often responding humorously or dismissively. To address this, we adopted an indirect approach to gender attribution for the main study, inspired by [20], especially for non-creature-like robots such as drones, where direct gender assignment poses challenges.

Based on the insights from the pilot study of our research, we propose the following hypotheses to guide our investigation:

***Hypothesis 1*** is based on the societal stereotype linking masculinity with technological robustness and utility, as outlined by gender schema theory [37]. This theory suggests that cultural stereotypes, such as associating strength and functionality with masculinity, influence our perceptions, leading us to view the Spot robot as more masculine compared to the Mini-Cheetah robot.

***Hypothesis 2***, grounded in Anthropomorphism [38] and gender schema theories, posits that the physical and behavioral attributes of non-humanoid robots influence the gender perceptions assigned by humans. Attributes that align with stereotypical masculine or feminine traits are likely to lead to corresponding gender attributions.

***Hypothesis 3*** draws on social role theory [39], anticipating that perceived gender affects robots' role assignments due to societal norms around gender roles, suggesting that tasks seen as more masculine or feminine will influence role assignment to robots.

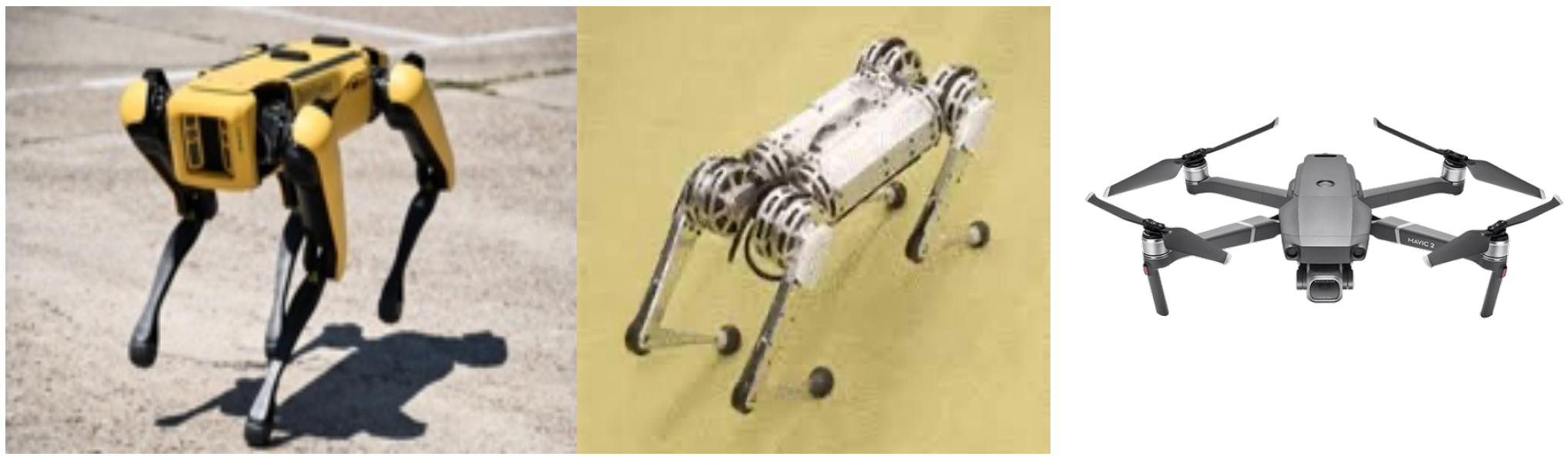

**Figure 1 A selection of robotic stimuli used in Survey 1. From left to right: Spot, Mini-Cheetah, and the drone.**

### B. *Survey Detail*

The survey began by asking participants to assign a name to the Spot and Mini-Cheetah robots. This provided insight into how participants perceived the robots, either as alive creatures, human-like entities, or mechanical objects. This approach to naming was inspired from [40], underscoring its efficacy in probing how individuals perceive and designate non-human entities. To analyze the assigned names, we inspired from classification categories as outlined in [38]. The names were sorted into two primary groups: anthropomorphic and non-anthropomorphic. Within the anthropomorphic category, names were further subdivided into male, female, or both-gender-associated names. The non-anthropomorphic category features three subcategories: animal-kind, machine-kind, and things-kind. These subcategories were further dissected into male, female, and neutral classifications for both animal-kind and machine-kind. The analysis entailed utilizing dictionaries and engaging five independent raters to evaluate the names. The evaluators rated each name, and the results were aggregated to determine common usage and associations. This approach allowed for an unbiased understanding of the naming patterns.

Then, participants rating specific attributes associated with robots. To accomplish this, we carefully selected 20 adjectives from [41], encompassing traits conventionally associated with male and female gender attributes. The chosen adjectives included ten behavioral characteristics and ten physical attributes. Participants were then asked to assign a rating to each adjective using a scale of 1 to 5. Table I lists the selected attributes for our study.

Subsequently, participants evaluated the suitability of 10 distinct occupations for the robots under review, based on gender categories proposed by [42]. They assigned ratings from 1 to 5 for each occupation, indicating their trust level for each robot's performance in these roles. Higher scores indicated a greater perceived alignment between the robots and the designated occupations. Table II lists the occupations traditionally viewed as male or female.

TABLE I THE PHYSICAL AND BEHAVIORAL ATTRIBUTES.

| Gender | Behavioral Attributes | Physical Attributes |
|---|---|---|
| **Males** | Assertive, Aggressive, Authoritative, Tough, Strong | Athletic, Heavy, Angular, Broad Shoulders, Rugged |
| **Females** | Empathetic, Delicate, Friendly, Sensitive, Compassionate | Graceful, Sleek, Slender, Elegant, Smooth |

Furthermore, to explore participants' perceptions of gender associations and biases concerning robots, we introduced a ranking system for the selected occupations linked to each robot. Participants were asked to rank each occupation based on which they believed the robot would perform most efficiently in. The occupations encompassed roles traditionally associated with males, females, and those considered gender-neutral, such as security guard, health care assistant, and food server.

TABLE II TRADITIONALLY MALE AND FEMALE OCCUPATIONS.

| Gender | Occupation |
|---|---|
| **Males** | Police Officer, Firefighter, Construction Worker, Miner, Mechanics Assistance |
| **Females** | Nurse, Childcare, Housekeeper, Receptionist, Therapist |

Finally, participants were directly asked to indicate their perception of the gender of the robots. They were instructed to provide a rating on a scale of 1 to 5, where 1 represented a perception of the robot as more feminine, 5 as more masculine, and 3 as gender-neutral. Participants were also asked to provide personal information, including their age, race, and level of education.

### *C. Participants*

The main survey involving 150 participants comprised 81 males, and 69 females. The participants had a diverse range of racial backgrounds, with representation from Asian (10%), Middle Eastern (19.3%), European (32%), African (6.7%), and Latino (8%) backgrounds. The participant's age range of 18 to 60 years, with the means (M) of 32 and the standard deviation (SD) of 6.78. Before the study, participants were asked about their fluency in English, as the survey questions were presented in English.

### *D. Survey Instrument*

The study employed online methods to gather data. For the online questionnaire, participants accessed a web-based form where they were provided with a video showcasing the Spot and Mini-Cheetah robots and the commercial drone. The video included demonstrations of their capabilities, and participants were also shown photographs of the robots.

The study utilized different robot stimuli, including the Spot and Mini-Cheetah robots and the commercial drone (a and a DJI Mavic 2 drone). The corresponding photos of these robots are displayed in Fig. 1.

### *E. Results*

The results indicate that the Spot robot was predominantly perceived as masculine by participants, with a mean score of 3.95, SD=1.1. In contrast, the Mini-Cheetah robot was viewed as more feminine compared to the Spot robot, albeit with a more neutral mean score of 3.1, SD=1.3. The drone was also attributed masculine gender perceptions, receiving a mean score of 3.61, SD=0.9. A repeated-measures ANOVA confirmed significant differences in gender perception across the three robots ($F(2, 298) = 8.45$, $p = 0.0003$, partial $\eta^2 = 0.054$**)**, though the effect size indicated these differences were modest. Post hoc comparisons revealed that the Spot robot was perceived as significantly more masculine than the Mini-Cheetah ($p = 0.001$) and the Drone **(**$p = 0.005$**)**. However, the difference in gender perception between the Spot robot and the Drone was not statistically significant ($p = 0.07$).

In naming the Spot, most of the anthropomorphic names assigned were male (68.1%), followed by gender-neutral (21.8%), and female (10.1%). Within the non-anthropomorphic category, machine-kind names constituted 27.9%, animal-kind names 55.3%, and things-kind names 17.8%. These classifications further illuminate the anthropomorphic and non-anthropomorphic naming tendencies among participants.

For the Mini-Cheetah robot, neutral anthropomorphic names were the most frequent, comprising 30% of the 150. Male-associated names followed at 20%, and female-associated names at 15%. Non-anthropomorphic names comprised 35% of the total, with machine-like, animal-like, and object-like contributing 10%, 15%, and 10%, respectively. For both, 45 names (or 15% of the total names) were found to be inspired by media sources such as popular culture, movies, and literature.

Fig. 2 shows the mean score of the different attributes of robots. The Spot robot was perceived with the highest masculine attributes, both behaviorally (M = 3.98, SD=0.87) and physically (M = 4.04, SD=0.65), leading to an overall masculine average of 4.01. The Mini-Cheetah was associated more closely with feminine attributes, particularly in behavioral aspects (M = 3.92, SD=0.96), resulting in the highest overall feminine average of 3.85 among the robots. The drone presents a balanced mix of masculine and feminine perceptions, with slightly higher masculine behavioral scores (M = 3.68, SD=1.2) but a higher feminine physical average (M = 3.3 SD=0.98). Its overall averages are relatively balanced between masculine (2.92) and feminine (2.77) attributes. A correlation analysis demonstrated a positive relationship between perceived masculinity behavioral attribute and suitability for male-dominated roles ($r = 0.35$, $p = 0.002$), while perceived femininity was positively correlated with suitability for female-dominated roles ($r = 0.42$, $p < 0.001$).

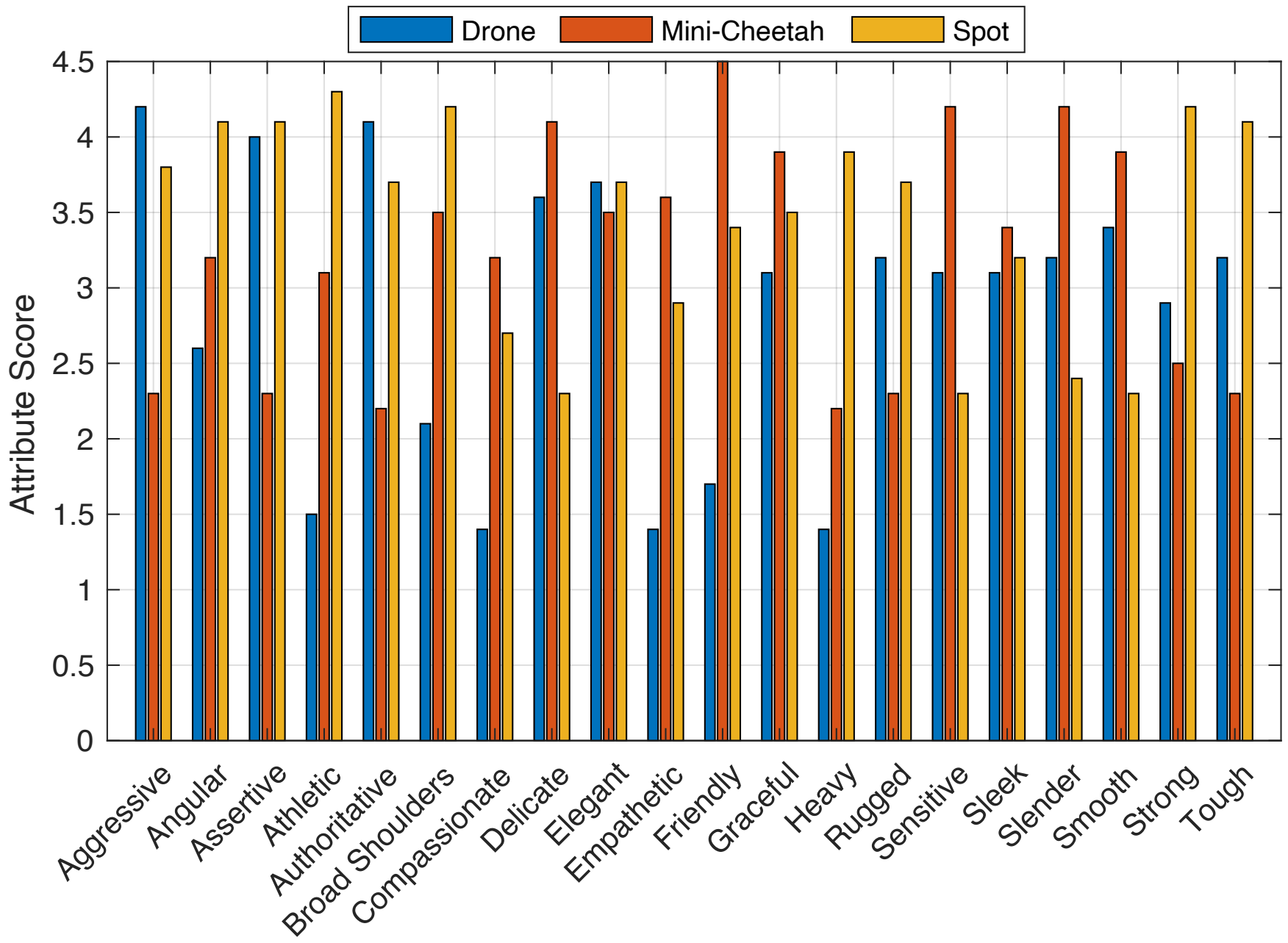

**Figure 2 The comparative attribute diagram.**

The Spot scored the highest in overall efficiency (M = 2.88) and higher in masculine occupations (M = 3.58), suggesting it is perceived as particularly suitable for roles traditionally associated with males. The Spot robot also scored lower in feminine occupations (M = 2.18), indicating a distinct gender-based perception of its suitability for various tasks. The Mini-Cheetah robot shows a more balanced profile, with its scores for feminine occupations (M = 2.78) being higher than for masculine occupations (M = 2.26). This suggests that participants might perceive the Mini-Cheetah robot as more versatile or suited for a broader range of roles, including those typically associated with females. The drone, with the lowest overall efficiency rating (M = 1.89), still scored moderately in masculine occupations (M = 2.32), but lower in feminine occupations (M =1.46). This may reflect a perception that drones, while versatile, are more suited to tasks that are not strongly associated with either gender but may lean slightly towards traditionally male roles. The result is shown in Fig. 3. A two-way repeated-measures ANOVA was conducted with Robot Type and Occupation Gender Category (Male-Dominated, Female-Dominated) as within-subject factors. The analysis revealed a main effect of robot type ($F(2, 298) = 12.56$, $p < 0.001$, $\eta^2 = 0.078$) and occupation gender category ($F(1, 149) = 45.67$, $p < 0.001$, $\eta^2 = 0.235$.), as well as an interaction effect between robot type and occupation gender category ($F(2, 298) = 2.95$, $p = 0.054$, $\eta^2 = 0.019$). The main effects indicate that both robot type and occupation gender category independently influence suitability ratings. The interaction effect is marginally non-significant ($p = 0.054$), it suggests that the difference in suitability ratings between male-dominated and female-dominated occupations does not significantly vary across robots. Post hoc analyses showed that the Spot Robot was rated higher for traditionally male occupations, such as police officer and firefighter compared to female occupations like nurse and childcare worker ($p < 0.001$). In contrast, the Mini-Cheetah Robot received higher suitability ratings for traditionally female occupations, such as nurse and childcare worker , over male occupations $p = 0.01$).

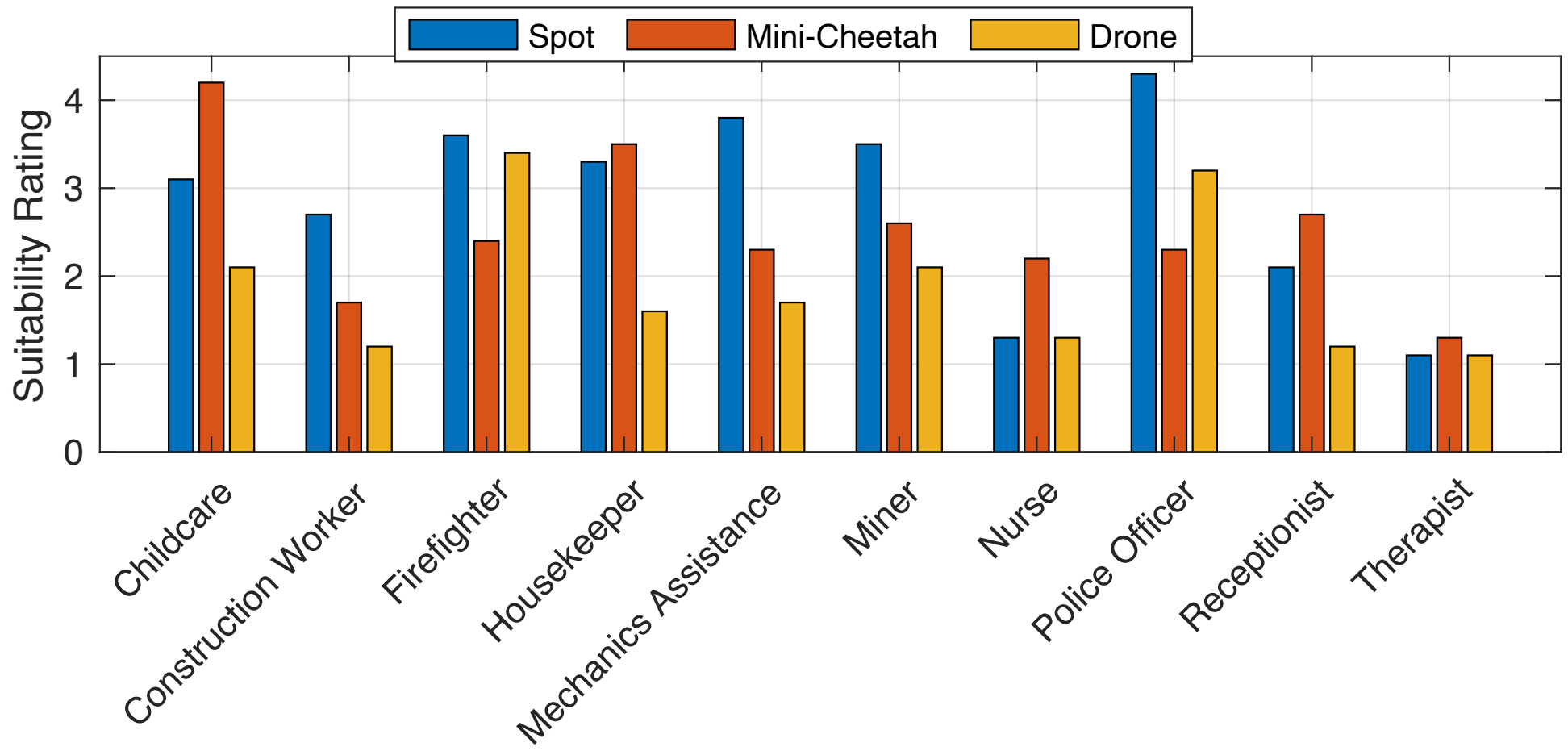

**Figure 3 The suitability ratings for the Spot, Mini-Cheetah, and the drone.**

In this survey, we employed a ranking for occupations. The Spot robot was predominantly ranked higher for security guard roles, while the Mini-Cheetah robot was considered more suitable for food delivery and healthcare assistance. As for the drone, it appears to excel in food delivery missions, which are considered neutral, compared to security and healthcare assistance, where its performance may be related to more neutral attributes people assign to it. Fig. 4 demonstrates the ranking scores of each robot for different occupations.

Our study corroborates the presupposed hypotheses and objectives, uncovering insights into gender perception within non-humanoid robotics. In alignment with Hypothesis 1, participants perceived the Spot robot as embodying masculine traits, which translated into a preference for the robot in traditionally male-dominated occupations, such as police and firefighting roles. This mirrors societal stereotypes that equate technological prowess with masculinity. The Mini-Cheetah robot, perceived with a feminine slant, was favored for roles traditionally associated with female attributes, like nursing and childcare, supporting Hypothesis 3 regarding gender-based role assignment.

Our data revealed a distinction between how behavior and physical appearance contribute to perceptions of gender. The drone, which scored higher in masculine behavioral attributes, was considered apt for roles that necessitate such traits, like police work. However, its lower scores in masculine physical attributes led to it being deemed less suitable for physically demanding occupations such as construction work. This divide highlights the challenge of creating robots for particular tasks and shows how important it is to think about both behavior and physical appearance when designing them.

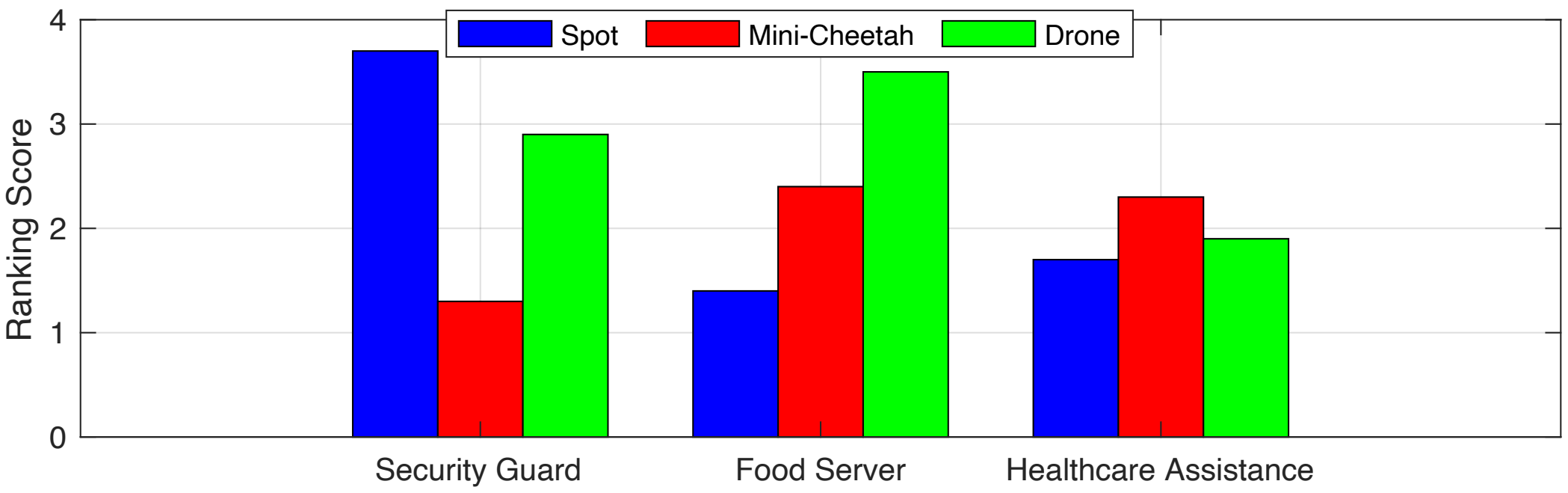

**Figure 4 The average ranking score for traditionally male, neutral, and female occupation.**

Moreover, the versatility observed in the Mini-Cheetah robot, capable of transcending gendered role boundaries, suggests a potential advantage in designing robots with androgynous features that can be broadly accepted across various roles. In contrast, the Spot robot's strong masculine image may restrict its perceived utility, pointing to an opportunity for design enhancements that could broaden its role adaptability.

Additionally, investigating the attributes and assigning roles shows us that masculine attributes have a greater influence on participants when assigning roles to non-humanoid robots, while non-muscular physical attributes make non-humanoid robots more neutral and suitable for a wider range of roles. However, overall, people tend to trust a non-humanoid robot with a more masculine physical attribute. Furthermore, people rate anthropomorphic behavioral attributes such as empathy and compassion lower for non-humanoid robots, and this tendency is more pronounced for drones.

## III. STUDY 2: EXAMINATION OF DIFFERENT LEVELS OF GENDER CUES ON NON-HUMANOID ROBOTS

This study investigates the effects of gender cues and anthropomorphism on HRI in non-humanoid robots. It seeks to elucidate how ascribing gender-related attributes and human-like qualities to these robots affects various HRI dimensions. We have redesigned the Spot robot, incorporating features that enhance its feminine, masculine, machine-like, and animal-like attributes. This modification aims to assess how varying levels of gender cues and anthropomorphism impact comfort and trust in role assignments and their overall influence on HRI dynamics.

Our study presents hypotheses based on HRI, to examine how gender cues in non-humanoid robots affect human perceptions and interactions:

***Hypothesis 1***: Gender stereotypes will influence participants' perceptions of a robot's task efficiency and their preference for robots as teammates, drawing on gender schema and social role theories.

***Hypothesis 2***: Anthropomorphic design will affect the perception of the robot's gender and increase user comfort, based on anthropomorphism theory.

***Hypothesis 3***: Gender cues in robot design, influenced by the Uncanny Valley theory [43], may affect user politeness towards robots, especially in response to errors.

### *A. Survey Instrument*

In this survey, advanced artificial intelligence-based generative models, DALL·E-2 [44], were utilized to redesign of the Spot robot. We instructed the model to generate diverse versions of it, including a more feminine, a masculine, a canine-shaped variant, and the original designs, illustrated in Fig. 5. In this survey, participants were randomly separated into four categories.

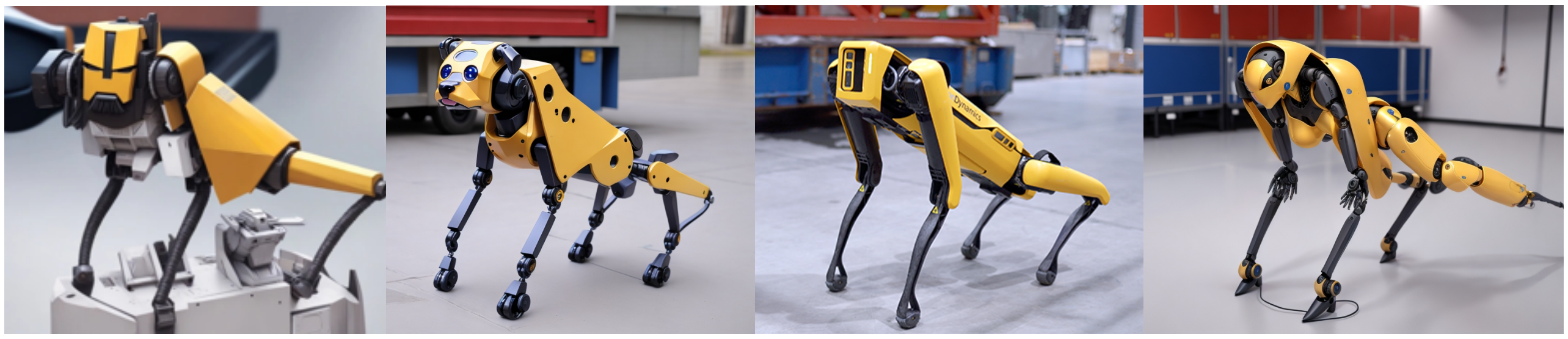

**Figure 5 Different designs of Spot generating by generative AI.**

For each design variant, a 10-second video was created in which the robot introduced itself against various backgrounds. A consistent dialogue was used across all versions:

*"I am Spot. I can assist you in various applications and possess numerous capabilities."*

The masculine design utilized a male voice, the feminine design employed a female voice, the dog-like design conveyed its message through barking accompanied by subtitles, and the original design featured a neutral-gender voice. All voiceovers were generated using Siri [45]. The reason behind this is that adding voice can add Anthropomorphic to robots [46].

### B. Pilot Study

In our study, we initiated with an online pilot study conducted in the University of Luxembourg involving 30 participants to assess gender perceptions of various voice samples and designs of the Spot robot using a 5-point Likert scale, where 1 signified a feminine perception and 5 indicated a masculine perception. For each design variation of the Spot, three designs were presented for evaluation regarding their perceived gender and realism, aiding in the selection of the optimal design for survey.

The analysis revealed that the neutral voice sample was slightly inclined towards masculinity (M = 3.2, SD = 1.2). Voice samples explicitly identified as male and female were rated accordingly, with means of 4.5 (SD = 0.4) and 1.4 (SD = 0.3), highlighting clear gender distinctions.

Within the Spot robot designs, the version intended to be masculine was perceived as masculine (M= 4.5, SD = 0.4). The design labeled as feminine was rated as less masculine (M = 1.3, SD = 0.6), and the dog-shaped variant was rated with a mean of 3.2 (SD = 0.8), indicating diverse perceptions of anthropomorphism and gender.

Furthermore, participants were asked to rate adjectives traditionally associated with male or female traits which use in the main survey. The low standard deviations observed suggest a broad agreement among participants, reflecting a conformity with established societal norms and gender stereotypes. These preliminary findings ensuring the methodological soundness and relevance of our investigation.

### C. Participants

We surveyed 120 participants from the University of Luxembourg aged 20-55 years. Among the participants, 56% identified as male and 44% as female, with a mean age of 32 and a standard deviation of 10.32. Given the diverse composition of the university's student body, the participants represented various racial backgrounds, including Asian (10%), Middle Eastern (26%), European (48%), Black or African American (6%), and Latino (10%). Regarding educational qualifications, we inquired about the participants' levels of education, revealing that 60% were either Ph.D. students or held higher degrees, 30% held master's degrees, and the remaining participants fell into other categories.

### D. Survey Detail

To assess participants' perceptions of the Spot's efficiency in performing various tasks, we asked them to evaluate the likelihood that the Spot robot could complete specific jobs. We presented participants with 10 occupations listed in Table V. Participants were instructed to rank each parameter on a scale from 1 to 5, with 1 indicating low likelihood and 5 indicating high likelihood of the Spot's success in that task.

**TABLE V OCCUPATION.**

| Category | Occupations |
|---|---|
| Traditionally Female | Nurse, Childcare, Housekeeper, Flight Attendant, Secretary |
| Traditionally Male | Construction Worker, Firefighter, Mechanic, Security Guard, Police |

Additionally, to examine the influence of gender cues on the selection of a robot as a teammate, we devised two scenarios aimed at evaluating participants' trust in the Spot robot for collaborative tasks. Trust was rated on a scale from 1 to 5, where 1 indicated minimal trust and 5 represented maximal trust. The first scenario was traditionally male-dominated, involving search and rescue operations to locate survivors. The second scenario was traditionally female-oriented, focusing on assistance in a cooking competition. Following these assessments, participants were asked to choose between the Spot robot and two humanoid robots (depicted in Fig. 6), determining their preferred teammate for each scenario. This comparison is due to the impact of non-humanoid and humanoid robots, regarding gender cues, on trust and acceptance.

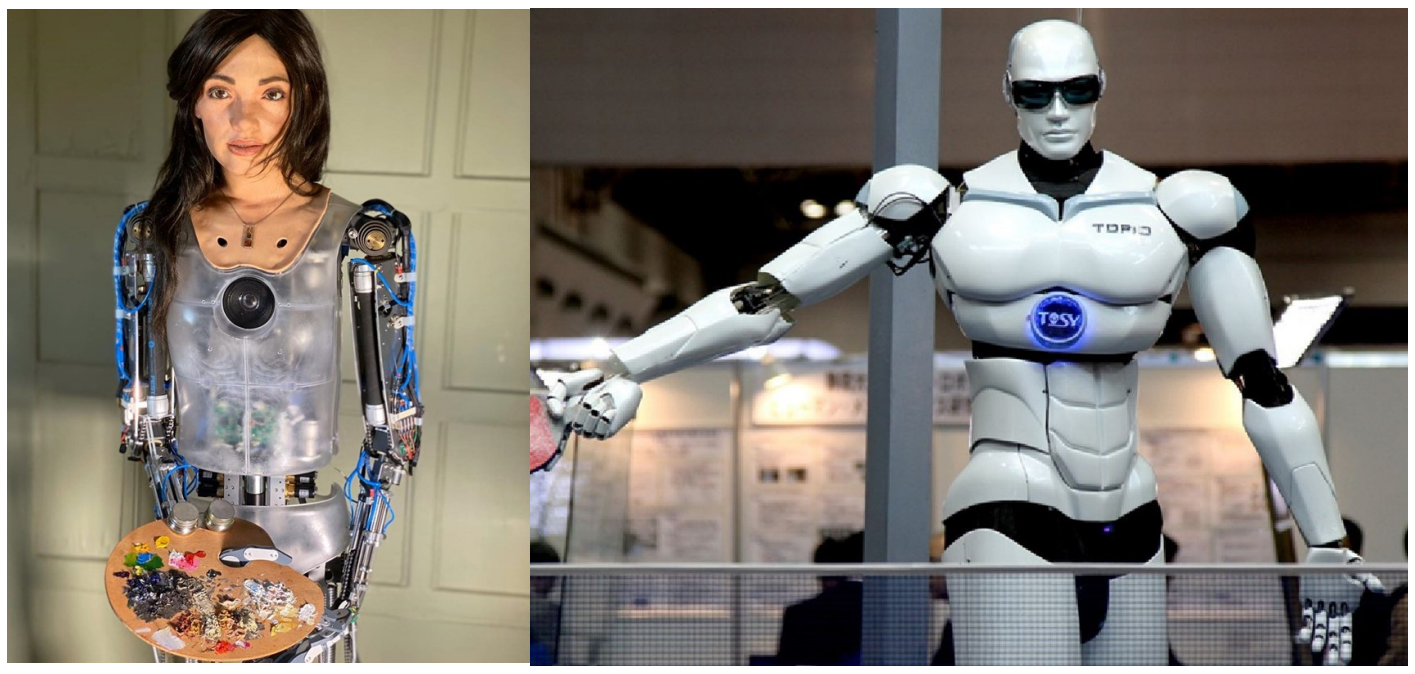

**Figure 6 The humanoid robots used in the survey 2.**

Furthermore, to assess the impact of gender cues on participants' comfort level with the Spot robot, we asked participants to rate their comfortability in being around the robot for an extended period on a scale from 1 to 5.

Then, we aimed to investigate whether different levels of gender cues in non-humanoid robots can influence participants' behavior towards the robots, specifically focusing on politeness. Participants were asked to rate their likelihood of exhibiting aggressive behavior towards the Spot robot if it made a mistake on a scale from 1 to 5. Finally, we directly inquired about the participants' perceptions of the robot's gender using a 1-5 Likert scale, where 1 represented the most feminine, 5 the most masculine, and 3 neutral.

### *E. Results*

As mentioned, the Spot was categorized into 4 distinct design categories, and each survey was conducted separately for each category to prevent design bias from influencing participants' responses. These categories included Spot A, featuring a feminine design with a female voice; Spot B, characterized by a masculine design with a male voice; Spot C, designed to resemble a canine shape with barking sounds and subtitles; Spot D, presenting an original design with a neutral voice.

The analysis revealed a statistically effect of gender cues on the gender perception of participants across four design variants, it evidenced by a one-way ANOVA [$F(3, 116) = 4.22$, $p = 0.007$], indicating that design elements influence gender attribution to robots. Further exploration through post hoc analyses using the Tukey HSD test indicated a difference in gender perception based on the design cues. The Spot B recorded a higher mean score ($M = 4.4$, $SD = 0.3$) in gender perception compared to the Spot A ($M = 1.75$, $SD = 0.4$), highlighting a clear distinction in gender attribution by participants. Meanwhile, the designs intended to be gender-neutral, Spot C and Spot D, achieved mean scores ($M = 3.1$, $SD = 0.7$ and $M = 3.3$, $SD = 0.8$, respectively) that positioned them between the explicitly gendered designs.

A one-way ANOVA revealed an effect of gender cues on perceived efficiency in task performance [$F(3,116) = 4.56$, $p = 0.017$]. Post hoc comparisons indicated that the Spot B, with an average suitability score of 3.42, was perceived as the most appropriate across all occupations. Conversely, the Spot A, with the lowest overall average (3.01), was deemed the least preferred for general occupations. Spot C, excelling in specialized tasks, especially related to childcare, indicating specific designs can enhance trust in certain contexts. Fig. 7 demonstrate the average suitability score for each design for average traditionally male and female occupations. The study shows different trust to the different variant of Spot in gender-typical roles. Spot A was preferred for traditionally female roles. Spot B favored traditionally male, against traditionally female occupation. This variation highlights robots' versatile uses, shaped by societal standards and context, and shows how gender-related design cues affect trust in different occupations.

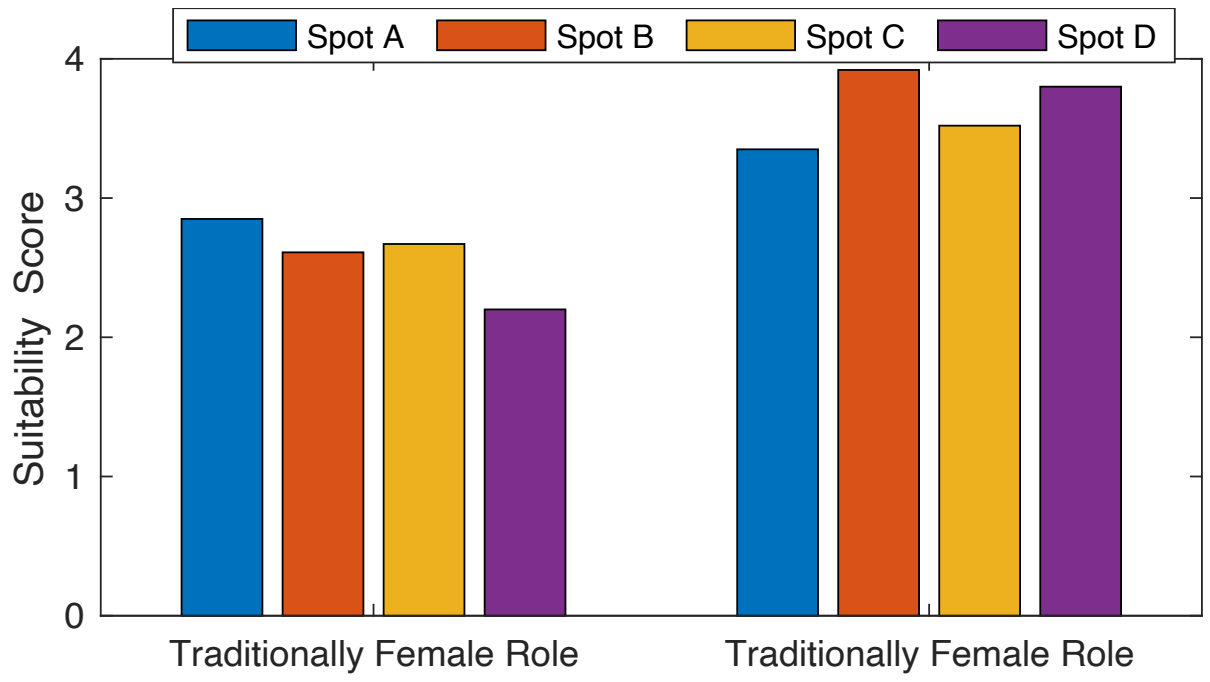

**Figure 7 Average suitability score regarding different designs of Spot for different occupations.**

For selecting Spot as a teammate, participants exhibited a preference for Spot B in the search and rescue in jungle, and following it they prefer Spot C and Spot D and least Spot A. This trend suggests a perception of Spot B as being more adept in challenging environments. In the cooking competition, Spot A was slightly more favored, reflecting societal norms that associate caregiving roles with femininity. During a search and rescue mission (traditionally male scenario) in the jungle, both male and female humanoid robots received lower scores compared to Spot B and Spot C. However, in the cooking competition, the Humanoid Female Robot scored higher than all Spot variants. The average scores for these results are shown in Fig. 8. Additionally, we explored a scenario in which the robot functioned as a nurse. The results indicated that people placed greater trust in humanoid robots over the Spot variants. However, the average trust score in this scenario was lower than in other scenarios, which were not as professionally oriented as this scenario.

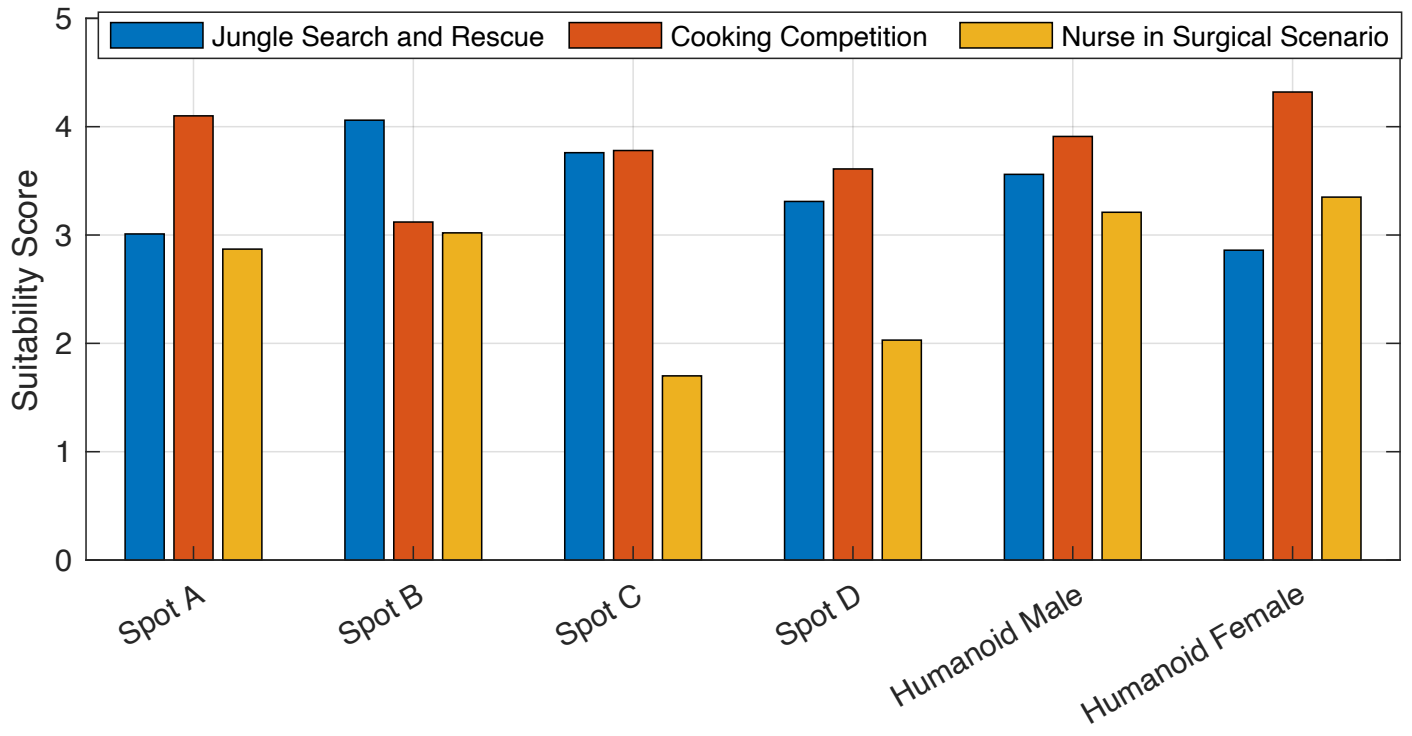

**Figure 8 Mean likelihood of choosing the different design of the spot for traditionally male, female, and neutral scenarios.**

There was an effect of gender cues on comfort level at the $p<0.05$ level for the four conditions [$F(3, 116) = 6.85$, $p = 0.034$]. Post hoc comparisons using the Tukey HSD test indicated that the mean score for the Spot A (M=4.2, SD=0.5) was different than the Spot B (M=2.9 SD=0.8), Spot D (M=3.2, SD=0.7), and the Spot C (M=3.4, SD=0.4).

There was an effect of gender cues on politeness at the $p<0.05$ level for the four conditions [$F(3, 116) = 4.32$, $p = 0.016$]. Post hoc comparisons using the Tukey HSD test indicated that the mean score for the Spot A (M = 2.3, SD = 0.5) was different from the Spot B (M = 2.8, SD = 0.65), the Spot D (M = 3.2, SD = 0.7), and the Spot C (M = 2.1, SD = 0.9). It is shown that people tend to be less aggressive towards Spot C, which has a canine shape. However, adding gender cues is related to politeness, especially regarding Spot A, which has a feminine design.

Survey 2's findings substantiate the role of gender cues in HRI, as outlined by our hypotheses. ***Hypothesis 1*** was confirmed, with a pronounced preference for the masculine-designed Spot B, illustrating how gender stereotypes influence both the perceived efficiency of robots and their preferability as teammates. ***Hypothesis 2*** found validation in the increased comfort levels with the feminine-designed Spot A, demonstrating the positive impact of anthropomorphic designs with distinct gender cues on user comfort in HRI contexts. ***Hypothesis 3*** shows the effect of gender cues on people politeness, but on the other hand, it demonstrates that while people are less likely to be aggressive towards robots with added gender cues, robot designs that exhibit familiar features tend to further reduce aggressive behavior. This may be due to the animated face of Spot C.

The study also ventured into new territory with Spot C, a canine-shaped design perceived as neutral or specialized, pointing to the potential of non-traditional, non-anthropomorphic designs in fulfilling unique roles, a relatively unexplored domain in HRI. This investigation emphasizes the need to consider gender cues in robot design, not just for enhancing interaction dynamics but also for facilitating broader societal integration. Notably, the study highlights a distinct preference for non-humanoid robots like Spot in non-traditional human tasks, such as search and rescue operations, suggesting that non-humanoid forms may be better suited for certain tasks historically associated with non-human agents.

## IV. Study 3: Assessing the Impact of Gender Cues in non-humanoid robots Design on Role Assignment in Missions

This section explores the impact that gender cues and anthropomorphic designs have on human trust and acceptance in real world operational settings, particularly in SAR missions. The research is focused on understanding how varying levels of anthropomorphic design—including gender-specific vocalizations, movement behaviors, and visually gendered elements—affect human perceptions. Key areas of investigation include perceived safety, comfort levels, and the politeness of interactions with drones, which collectively influence how close humans are willing to interact with these machines. To achieve these objectives, the study manipulated drones' design features, such as size, color, and other elements traditionally associated with masculine or feminine traits. These design alterations were informed by existing research[47, 48], which discusses the application of gendered adjectives in design and their psychological impacts on human interaction. This approach allows for a detailed analysis of how subtle design cues can shift human expectations and operational efficiency in high-stakes environments like SAR operations.

### *A. Survey Instrument*

In this survey, artificial intelligence-based generative model, DALL·E-2 [44], was utilized to refine the design of Mavic-2 pro. These models were illustrated in Fig. 9, encompassing designs that were more distinctly feminine, masculine, and an animal-shaped (canine) variant, alongside the original configuration.

For each design variant, a 5-second video was created in which the drone introduced itself against various backgrounds. A consistent dialogue was used across all versions: "I am a drone. I can assist you in various applications and possess numerous capabilities." Voiceovers for each variant were distinct: the masculine version featured a male voice, the feminine version a female voice, the canine-shaped design communicated through barks with accompanying subtitles, and the original design used a gender-neutral voice. In addition, the original design was presented also, without voice narration, the latter including subtitles to convey its capabilities, which is "This is a DJI drone, it can assist you in various application and possess numerous capabilities." Participants in the study were randomly assigned to one of the five categories. The survey included video and images snippet for each drone variant to showcase the UAV's operational capabilities in post-disaster scenarios, featuring different speeds. Subsequently, they were asked to complete a survey, which on average took less than 5 minutes. It should be noted that the survey was conducted online.

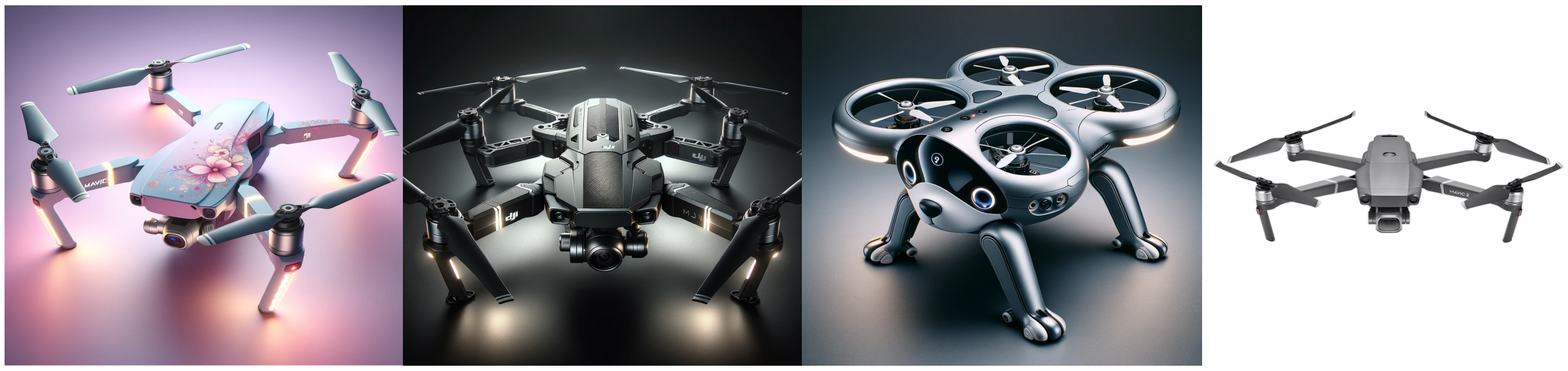

**Figure 9 The different variant of drone's design from left drone A, drone B, drone C, and drone D.**

### *B. Pilot Study*

Before the main investigation, an initial pilot study was conducted in person and involved participants within the age range of 25 to 55 years. Participants were asked to express the gender they attributed to a drone on a scale from 1 to 5, where 1 signified a stronger feminine perception. To validate the chosen attributes and occupations, the drones' design, and voice choice, participants rated each adjective based on its typical association with males or females. The results, with low standard deviations, indicated strong participant agreement. Mean ratings are generally aligned with societal norms and accepted gender stereotypes. Also, the participants were asked to rate the perceived gender of various voice samples, and drone designs for the efficiency of the survey instrument. The results validated the parameters.

### *C. Survey Detail*

In the survey, participants first evaluated various drone designs based on attributes associated with masculinity and femininity, as outlined in [41]. They then assessed the drones' suitability for ten specific occupations, evenly split between traditionally male and female roles [42]. Additionally, participants provided feedback on their comfort and perceived safety when in close proximity to each drone, including their willingness to allow the drones to land on their hands. To assess perceptions of the drones' effectiveness in search and rescue (SAR) missions, participants ranked the likelihood of successful mission completion across different scenarios. They also evaluated how drone speeds (5, 20, 50 km/h) influenced human proximity and comfort levels, along with hypothetical reactions of survivors when encountering drones post-disaster. Participants further indicated their willingness to seek help from drones in disaster situations and potential aggressive responses if the drones experienced operational errors. It's important to note that all responses were measured using a Likert scale [49], ranging from 1 (lowest) to 10 (highest). Finally, participants rated the perceived gender of the drones on a scale from 1 (feminine) to 10 (masculine), with 5 representing a gender-neutral perception. Demographic information, including age and gender, was also collected.

### *D. Results*

The survey included 150 participants, with 56% male and 44% female, aged 18-60, representing diverse backgrounds: Asian (20.6%), Middle Eastern (39.4%), European (28.6%), African (5.4%), and Latino (6%). English fluency was confirmed for the English-language survey.

The drones were organized into five unique design groups, with each group being surveyed independently to avoid design bias affecting the participants' responses. The categories were as follows: Drone A, which had a feminine design and a female voice; Drone B, which featured a masculine design and a male voice; Drone C, designed to mimic a canine form with barking sounds and subtitles; Drone D1, which offered an original design with a neutral voice; and Drone D2, which displayed the same original design but without a voice (refer to Fig. 1). The findings revealed that participants perceived Drone A as distinctly feminine with a Mean Score (MS) of 3.51 and a Standard Deviation (SD) of 2.58. Drone B was identified as notably masculine, registering a MS of 8.33 and an SD of 1.83. Drone C was seen as neutral with a MS of 5.58 and an SD of 1.25. Similarly, Drone D1 was also viewed as neutral, scoring a MS of 5.88 and an SD of 2.05. In contrast, Drone D2 was perceived as more masculine than neutral, with a MS of 6.5 and an SD of 1.05. This suggests that even subtle design changes, like the presence or absence of voice, can shift gender perceptions.

Fig. 10 presents the average scores for various robot attributes, covering five characteristics traditionally linked to both male and female traits. The findings indicate that Drone A obtained a mean score of 5.738 for Traditionally Female (TF) attributes and 4.45 for Traditionally Male (TM) attributes. Drone B received a mean score of 4.56 for TF attributes and 6.7 for TM attributes, while Drone C received mean scores of 5.434 for TF and 5.1 for TM attributes. Similarly, Drone D1 received mean scores of 4.36 for TF attributes and 5.153 for TM attributes, and Drone D2 had a mean score of 4.85 for TF attributes and 5.55 for TM attributes. The results suggest that Drone B, characterized by male attributes, excels in masculine attributes, whereas Drone A, embodying feminine attributes, demonstrates higher mean scores for female attributes. These findings underscore that gender cues can influence not only perceptions but also the assignment of specific qualities to non-humanoid robots.

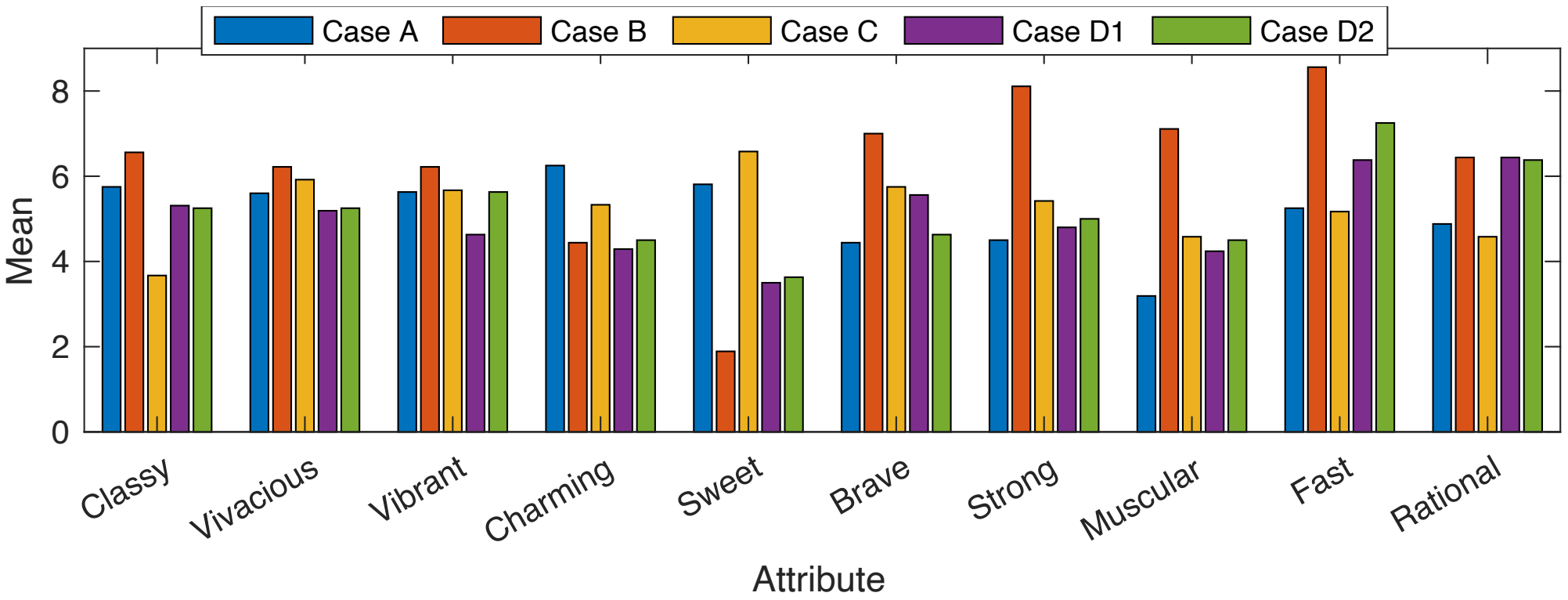

**Figure 10 The TM and TF attributes. The first five attributes are TF, while the others are TM.**

Fig. 3 illustrates trust levels in traditionally gendered occupations by drone, featuring four male-dominated roles and the remaining female-dominated roles, as validated in a pilot study. Drone A averages a 6.03 trust level, B scores 6.84, C 6.46, D1 6.12, and D2 6.61. The data suggest masculine design cues in drones correlate with higher trust in task efficiency, with Drone B leading in perceived efficiency across all roles, especially those considered masculine, and D1 following closely. Drone C is noted for excelling in specialized tasks due to its design, indicating specific designs can enhance trust in certain contexts.

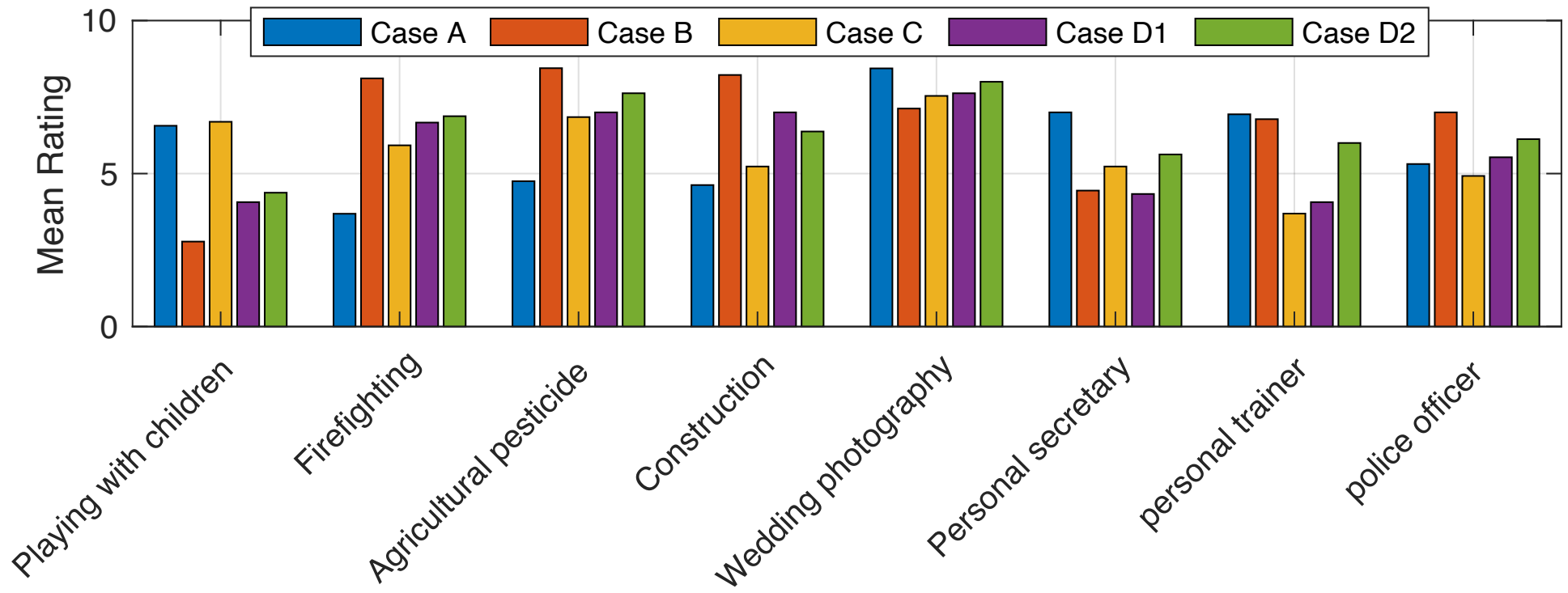

**Figure 11 Trust for each drone design in TM and TF occupations.**

The study shows different reactions to drone use in gender-typical roles and SAR tasks. Drone A was preferred for TF roles, scoring 7.125, over SAR (5.71875) and TM roles (5.0625). Drone B favored TM and SAR uses, with scores of 7.7111 and 8.5556, against TF's 5.2917, indicating a preference for more demanding or urgent uses. SAR was well-received across drones, especially Drone B (8.5556) and Drone D2 (7.5625). This variation highlights drones' versatile uses, shaped by societal standards and context, and shows how gender-related design cues affect trust in TF and TM occupations.

As shown in Fig. 12, participants exhibited the highest comfort level interacting with the Drone A, with an MS of 7.125, and were most willing to allow the drone to land on their hand, indicating a level of trust in its safety and functionality. Drone B showed a slightly higher inclination to consider purchasing (4.875) and paying more, reflecting trust in its efficiency, though participants were more ambivalent about interaction comfort. Drone B also displayed the highest pricing agreement mean (6.6667), suggesting recognition of its value despite safety concerns, indicated by lower safety feeling ratings (4.8889). Drones C, D1, and D2 presented mixed but generally moderate views on interaction comfort and likelihood of drone purchase. Participants in Drone D1 were notably less inclined to consider purchasing the drone (3.9333) and expressed the highest likelihood of feeling angry if it malfunctioned (6.0), indicating frustration. Safety perceptions and willingness to allow the drone to land on one's hand varied, reflecting ongoing reservations about drone proximity and interaction safety. Data suggests that the D1 neutral configuration is most likely to induce anger and provoke negative behavior, highlighting the influence of gender cues in design and people's willingness to accept them.

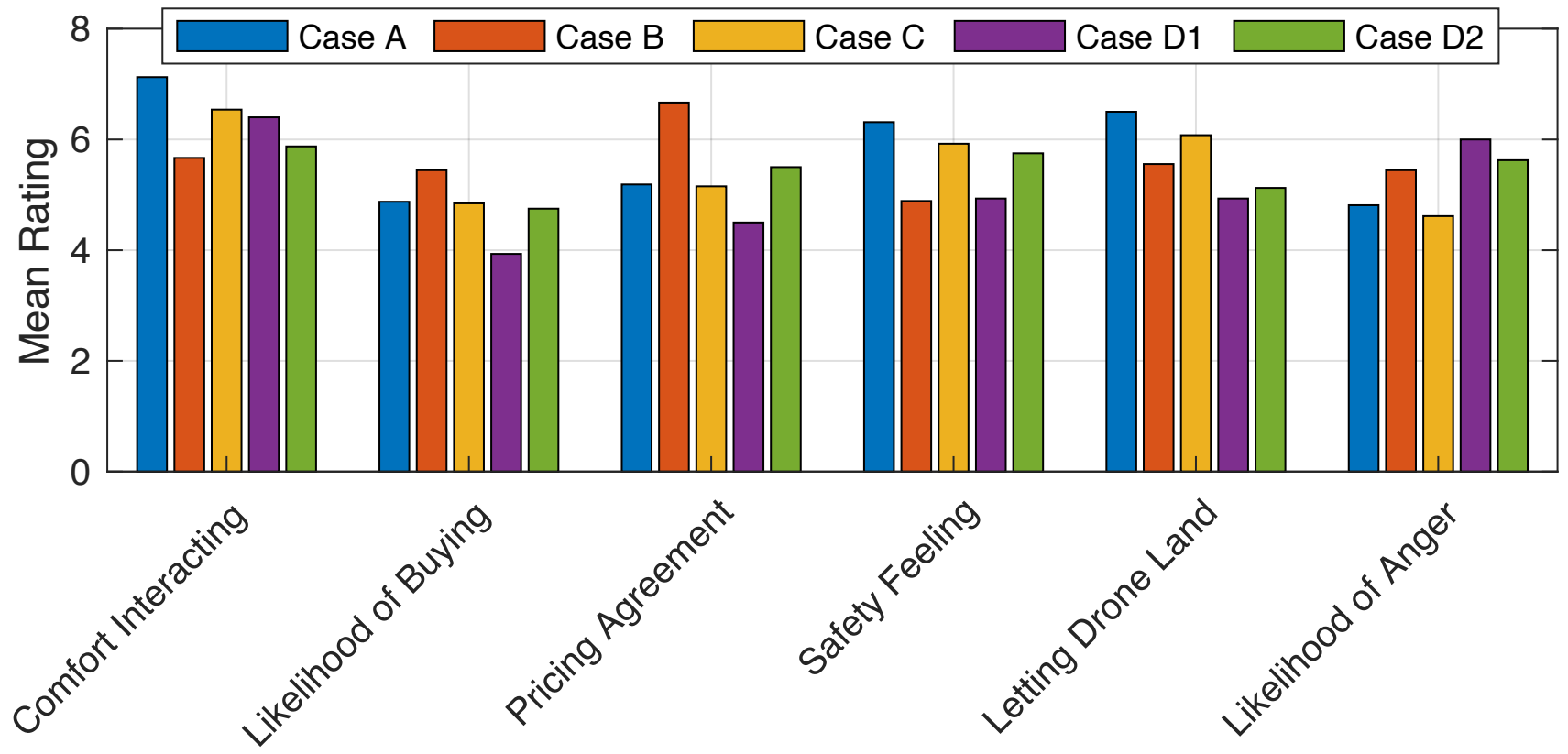

**Figure 12 Human acceptance based on different drone designs.**

As illustrated in Fig. 13, the assessment of people's willingness to use drones for SAR missions reveals intriguing insights into the perceived gender cues of the drones and their impact on participant responses. Case B, associated with more traditionally masculine gender cues, stands out as the scenario where participants expressed the highest likelihood of using the drone as a teammate in SAR competitions, with a mean rating of 8.6667. This contrasts with perceptions towards drones in Drones A and D1, where the MS are 6.375 and 6.875, respectively. Notably, while Drone B is considered to have more masculine cues, it still received a higher willingness rating to be used as a pilot (7.45) compared to other cases. Furthermore, when examining the likelihood of purchasing the drone for disaster assistance, Drones B, C, and D1 are among the highest, with MS of 6.7778, 6.7692, and 6.0667, respectively, showcasing a close range of responses suggesting that a more masculine perception could positively influence the decision to purchase drones for use in SAR missions. Specifically, Drone B not only rates highest in terms of being used as a teammate (8.6667) but also shows an openness to being piloted in real SAR missions (7.4444) and a robust willingness to purchase for disaster scenarios (6.7778).

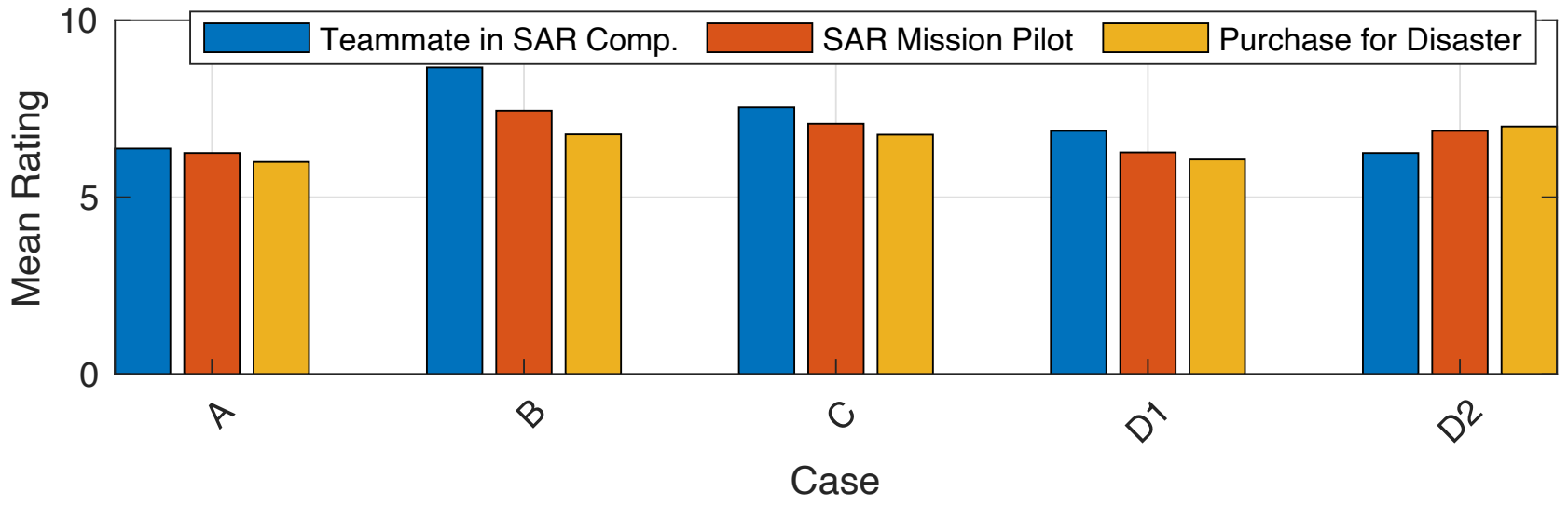

**Figure 13 The willingness to engage with different designs in SAR missions.**

These findings underscore a relationship between the perceived gender cues of drones and their acceptance for assigned roles. The data suggests that drones associated with more masculine traits, as reflected in Case B, tend to inspire greater confidence in their capabilities for teamwork and piloting in SAR contexts, as well as a higher propensity among participants to consider purchasing them for disaster assistance.

The analysis presented in Fig. 14 elucidates the relationship between drone speed and the likelihood of individuals running away when the drone approaches near participants at varying speeds. The results underscore a clear trend: as the drone's speed increases, so does the likelihood of individuals choosing to run, highlighting the impact of drone velocity on human behavior during high-risk missions. This observation is crucial, emphasizing that even when drone assistance is vital, the approach speed can influence the receptiveness and comfort levels of those being aided.

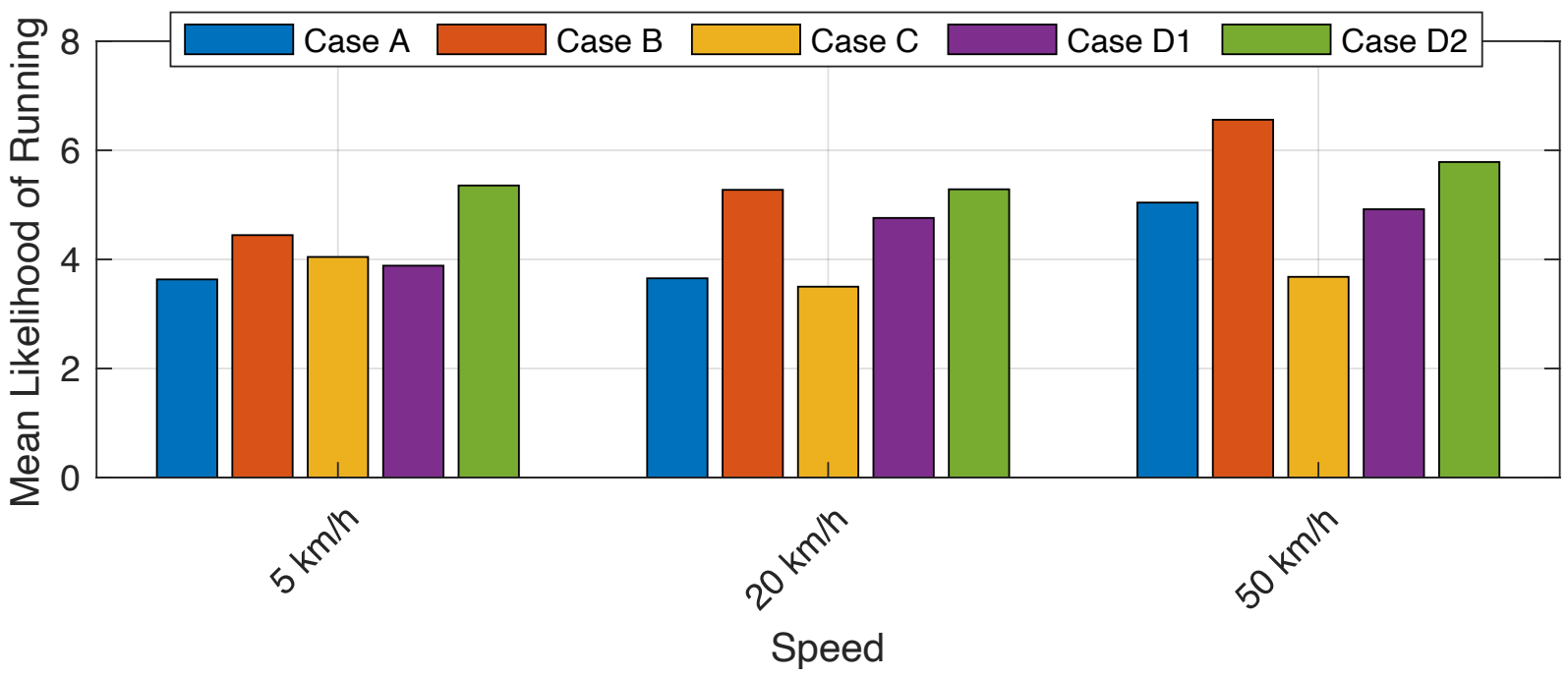

Figure 14 Likelihood of humans fleeing from different drone designs approaching at varying speeds.

Notably, Drone C, characterized by an animal-inspired design, demonstrates a lower likelihood of participants fleeing from it, evoking a sense of friendliness and approachability. Similarly, Drone A, which incorporates feminine design cues, shows a trend where participants are more tolerant and receptive, suggesting that sympathetic or relatable designs can reduce fear and enhance acceptance in human-drone interactions. Additionally, the neutral voice in Drone D1 appears to decrease the likelihood of participants fleeing compared to the voice-less D2, indicating that auditory feedback can subtly influence comfort levels and perceptions of approachability.

Participants generally perceived Drone A as more comforting, with a greater tolerance for closer proximity at lower speeds. In contrast, Drone B's masculine cues were more likely to be associated with participants running at higher speeds, possibly reflecting a subconscious link between masculine features and perceived aggression. However, this effect was less evident at lower speeds, suggesting that controlled operational settings may mitigate any intimidation.

The dog-shaped drone was particularly effective in reducing flight responses across varying speeds, highlighting the positive impact of approachable, non-humanoid designs on human-drone interactions. While speed remains a critical factor—participants preferred greater distances as drone speed increased—design elements that evoke familiarity and relatability can help balance operational efficiency with user comfort, especially in sensitive scenarios.

Overall, the findings emphasize that appearance plays a crucial role in human-drone interaction, with designs perceived as sympathetic or gentle tending to reduce fear and improve acceptance. Importantly, female participants were more likely than males to assign gender to drones, with males often perceiving them as neutral. This difference underscores the importance of considering diverse user perspectives in the design of non-humanoid robots.

## V. Discussion

This research set out to explore the dynamics of gender attribution in non-humanoid robots and its subsequent effects on HRI. Through a series of three comprehensive surveys, we investigated how design elements and gender cues influence human perceptions, trust, comfort, and role assignment concerning non-humanoid robots.

Survey 1 revealed a clear tendency among participants to attribute gender to non-humanoid robots based on their physical and behavioral attributes. The Spot robot was perceived as masculine, receiving male-associated names and being assigned to traditionally male roles such as police officer and firefighter. In contrast, the Mini-Cheetah robot exhibited a more balanced gender perception, slightly leaning towards feminine attributes, and was preferred for traditionally female roles like nursing and childcare. These findings align with Gender Schema Theory [34], which posits that cultural stereotypes and societal norms shape cognitive structures related to gender. The association of technological robustness and utility with masculinity influenced participants' perceptions, leading them to assign masculine traits to robots like Spot. The naming patterns further corroborate this tendency, with a number of participants choosing male-associated names for the Spot robot, while the Mini-Cheetah exhibited a more balanced naming distribution with a slight female inclination. Moreover, the attribution of roles based on physical and behavioral attributes underscores the influence of societal stereotypes in HRI. Participants linked physical attributes to the robot's capability to perform tasks efficiently, reflecting "performance-based trust" [50], while behavioral attributes were associated with roles requiring trust and collaboration, indicative of "relational-based trust"[51, 52]. For instance, the drone, despite its higher feminine physical average, was assigned to roles based more on behavioral cues, such as police officer, emphasizing the importance of behavioral attributes in trust-building.

Survey 2 demonstrated that explicit gender cues in robot design affect gender attribution and HRI outcomes. By redesigning the Spot we observed shifts in participants' perceptions and interactions. The masculine design was consistently rated higher for efficiency in traditionally male-dominated occupations, reinforcing the association between masculinity and task performance in high-stakes roles. Participants preferred Spot B as a teammate in challenging environments, reflecting trust in its perceived robustness. This supports Hypothesis and aligns with social role theory [53].

Conversely, the feminine design excelled in roles linked to caregiving and support, such as healthcare assistance and cooking competitions. Participants reported higher comfort levels with Spot A, indicating that anthropomorphic designs with feminine cues can enhance user comfort and engagement. This finding supports Hypothesis and is consistent with anthropomorphism theory [54], which suggests that attributing human-like qualities to non-human entities can improve relatability and interaction quality. Spot C's success in specialized tasks like childcare suggests that non-traditional designs can effectively enhance trust and suitability in specific contexts. This expands the scope of robot design by demonstrating that robots need not mimic human forms to be effective; familiar and friendly designs can also foster positive HRI.

Survey 3 focused on the impact of gender cues in non-humanoid robot design on role assignment in missions, particularly in SAR operations. By manipulating drones' design features we assessed how these modifications affect human perceptions of safety, comfort, and willingness to interact closely with drones. Participants perceived Drone B, designed with masculine cues, as more capable and trustworthy for high-stakes tasks. It received higher ratings for task efficiency, trustworthiness, and willingness to be used as a teammate in SAR missions. This supports Hypothesis 3, indicating that gender perceptions influence role assignment in critical environments.

Drone A, with feminine design cues, was favored in roles requiring approachability and comfort. Participants exhibited higher comfort levels and were more willing to allow Drone A to land on their hands, suggesting that feminine cues can enhance perceptions of safety and gentleness. An important finding was the effect of drone speed on human reactions. Higher speeds increased participants' likelihood of avoidance behaviors, regardless of design. However, Drone C, featuring an animal-inspired design, successfully reduced flight responses even at higher speeds. This suggests that incorporating non-threatening elements can improve user receptiveness, emphasizing the need to balance operational efficiency with user comfort.

Overall, Our study builds on the understanding that trust in robots is shaped by more than just their technical performance and reliability [8]. While performance remains important, our findings suggest that design features, including social cues and anthropomorphic elements, contribute to perceived trustworthiness and user acceptance. This observation aligns with emerging research [55]that emphasizes the role of design in enhancing human-robot interaction.

We contribute to the literature on human-robot collaboration by highlighting the importance of a robot's social presence—its ability to respond to social cues—in facilitating effective teamwork. Prior studies [8] have acknowledged the role of social cues, and our findings provide empirical evidence that these cues not only support task outcomes but also enrich collaboration through more engaging interactions. For instance, elements like voice modulation and aesthetic design contribute to building trust, demonstrating that well-considered design choices can enhance both adaptability and perceived reliability by Pinney, et al. [56].

Our results also indicate that trust varies depending on the perceived criticality of the robot's role and the social norms associated with those roles. In high-stakes tasks, stronger social biases and expectations appear to influence trust and interaction dynamics, while in less critical roles, these biases are less pronounced. This shows that social expectations and role criticality intersect in shaping how robots are trusted and accepted.

Moreover, our study refines how specific design features—such as contextual adaptability and behavioral traits—impact user perceptions and interaction outcomes. Previous research often viewed HRI outcomes primarily through the lens of functionality or anthropomorphism [57]. However, our findings suggest that subtle design elements, like emphasizing behavioral attributes (e.g., assertiveness or friendliness) or aligning with social norms, can positively influence cooperation, satisfaction, and perceived reliability among human users. Our exploration of non-anthropomorphic designs, such as a canine-shaped robot, highlights the potential of alternative forms in fulfilling specific roles. The preference for these designs in certain contexts suggests that effective HRI design does not rely solely on human-like features but can explore other forms that achieve both social integration and functional efficiency. In non-humanoid robots, integrating gender-specific behavioral cues—such as voice modulation, movement patterns, and task-related behaviors—is crucial for influencing user perceptions and enhancing acceptance in designated roles. Our findings indicate that behavioral attributes can independently drive role assignments, suggesting that designers should prioritize behavior modulation alongside or even over physical design alterations.

The study highlighted gender differences in gender attribution, with female participants more likely to assign gender to drones compared to male participants, who often perceived drones as neutral. This disparity underscores the necessity for inclusive design approaches that account for diverse user perceptions and reduce gender biases in robot design [31]. The study underscores the influence of cultural and societal norms on gender attribution and role assignment in HRI. Given the diverse participant demographics, the findings suggest that global robot design strategies must account for varying cultural perceptions of gender to ensure broad societal acceptance.

The preference for Drone B in SAR missions and the higher willingness to purchase gendered drones for disaster assistance indicate that gender cues can influence not only user trust but also market acceptance. Designers should consider the specific operational contexts and user demographics when incorporating gendered elements to optimize both functionality and acceptance.

The findings highlight the delicate balance between anthropomorphic design and functional utility. While gendered cues can enhance relatability and trust in specific contexts, overly rigid gender associations may limit the versatility of robots across diverse roles. Designers must navigate this balance to create robots that are both functionally effective and socially engaging without falling into the "uncanny valley" [11].

While our study primarily focused on binary gender categories (male and female), we extended our analysis by treating gender as a continuous scale ranging from 1 (most feminine) to 5 (most masculine). This approach acknowledges the fluidity and spectrum of gender identities beyond the traditional binary framework. By utilizing a 1-5 scale, we captured more nuanced perceptions of gender attribution, allowing participants to express degrees of gender association with each robot design. This method provided a more comprehensive understanding of how gender cues influence HRI, accommodating diverse user perspectives and reducing the limitations inherent in binary classifications.

### *1) Limitations and Future Research*

**Participant Demographics and Cultural Contexts**: Our study focused on binary gender categories of participants and did not fully explore the influence of cultural and racial differences background of participants on robot perception. Future studies should take a more inclusive approach to understanding how gender, culture, and ability intersect in HRI.

**Inclusion of Diverse Populations**: The study did not account for individuals with visual or auditory disabilities, and the sample was primarily composed of adults. Expanding future research to include a broader range of populations will provide a more comprehensive view of human-robot interactions.

**Real-World Application and Generalizability**: The generalizability of our findings should be tested in varied real-world settings.

**Exploring Non-Binary and Fluid Gender Representations**: Future research should investigate non-binary and fluid gender representations in robot design to enhance inclusivity and accommodate diverse user identities and preferences.

**Integration of Dynamic Behavioral Traits:** Exploring how dynamic behavioral traits—such as adaptability and responsiveness—interact with static design cues can lead to the development of more sophisticated and contextually responsive robotic systems.

## VI. CONCLUSION

Our work explores how the application of gender cues and anthropomorphic design influences perceptions of non-humanoid robots in terms of task performance and comfort for users in social interactions. The findings highlight that robot trust and comfort-considering varied user expectations-lie at the heart of effective HRI. Non-humanoid robots can decrease the gap between operational effectiveness and social acceptance by embedding design features that consider sociability standards.

This paper also shows the flexibility with which social roles and gender norms can be expressed through design, hence giving new insights into traditional HRI theories. It further outlines that societal biases need to be addressed in the design of robots, supporting inclusive approaches that balance task efficiency with social acceptability. Future HRI design should integrate functionality and sociability wherein robots play their intended roles while maintaining varied human expectations.

## Acknowledgement

This work was partially funded by the Fonds National de la Recherche of Luxembourg (FNR) under the project C22/IS/17387634/DEUS.

During the preparation of this work, the authors used ChatGPT in order to assist enhance grammar and readability. After using this tool/service, the authors reviewed and edited the content as needed and take full responsibility for the content of the publication.